\documentclass[letterpaper]{article} 
\usepackage{aaai25}
\usepackage{times}  
\usepackage{helvet}  
\usepackage{courier}  
\usepackage[hyphens]{url}  
\usepackage{graphicx} 
\usepackage{natbib}  
\usepackage{caption} 
\usepackage{algorithm}
\usepackage{algorithmic}
\usepackage{booktabs}
\usepackage{times}
\usepackage{amsmath}
\usepackage{inconsolata}
\usepackage{amssymb}
\usepackage{graphicx}
\usepackage{array}   
\usepackage{multirow}
\usepackage{makecell}
\usepackage{rotating}
\usepackage{caption}
\usepackage{subcaption}
\usepackage{enumitem}
\usepackage{bm}
\usepackage[bottom,hang,flushmargin]{footmisc}

\usepackage{newfloat}
\usepackage{listings}
\DeclareCaptionStyle{ruled}{labelfont=normalfont,labelsep=colon,strut=off} 
\floatstyle{ruled}
\newfloat{listing}{tb}{lst}{}
\floatname{listing}{Listing}
\title{Channel Merging: Preserving Specialization for Merged Experts}
\author{
    Mingyang Zhang\textsuperscript{\rm 1},
    Jing Liu\textsuperscript{\rm 2}, 
    Ganggui Ding\textsuperscript{\rm 1}, 
    Linlin Ou\textsuperscript{\rm 3}, 
    Xinyi Yu\textsuperscript{\rm 3}, \\
    Bohan Zhuang\textsuperscript{\rm 1}\thanks{BZ is the corresponding author.} 
}
\affiliations{
    \textsuperscript{\rm 1} Zhejiang University\\
    \textsuperscript{\rm 2} Monash University \\
    \textsuperscript{\rm 3} Zhejiang University of Technology

}

\usepackage{bibentry}

\begin{document}

\maketitle

\begin{abstract}
Lately, the practice of utilizing task-specific fine-tuning has been implemented to improve the performance of large language models (LLM) in subsequent tasks.
Through the integration of diverse LLMs, the overall competency of LLMs is significantly boosted. Nevertheless, traditional ensemble methods are notably memory-intensive, necessitating the simultaneous loading of all specialized models into GPU memory.
To address the inefficiency, model merging strategies have emerged, merging all LLMs into one model to reduce the memory footprint during inference. Despite these advances, model merging often leads to parameter conflicts and performance decline as the number of experts increases. Previous methods to mitigate these conflicts include post-pruning and partial merging. However, both approaches have limitations, particularly in terms of performance and storage efficiency when merged experts increase. To address these challenges, we introduce Channel Merging, a novel strategy designed to minimize parameter conflicts while enhancing storage efficiency. This method clusters and merges channel parameters based on their similarity to form several groups offline. By ensuring that only highly similar parameters are merged within each group, it significantly reduces parameter conflicts. During inference, we can instantly look up the expert parameters from the merged groups, preserving specialized knowledge. 
Our experiments demonstrate that Channel Merging consistently delivers high performance, matching unmerged models in tasks like English and Chinese reasoning, mathematical reasoning, and code generation. Moreover, it obtains results comparable to model ensemble with just 53\% parameters when used with a task-specific router.
\end{abstract}

%

\begin{figure*}[htbp]
    \centering
    \includegraphics[width=1\linewidth]{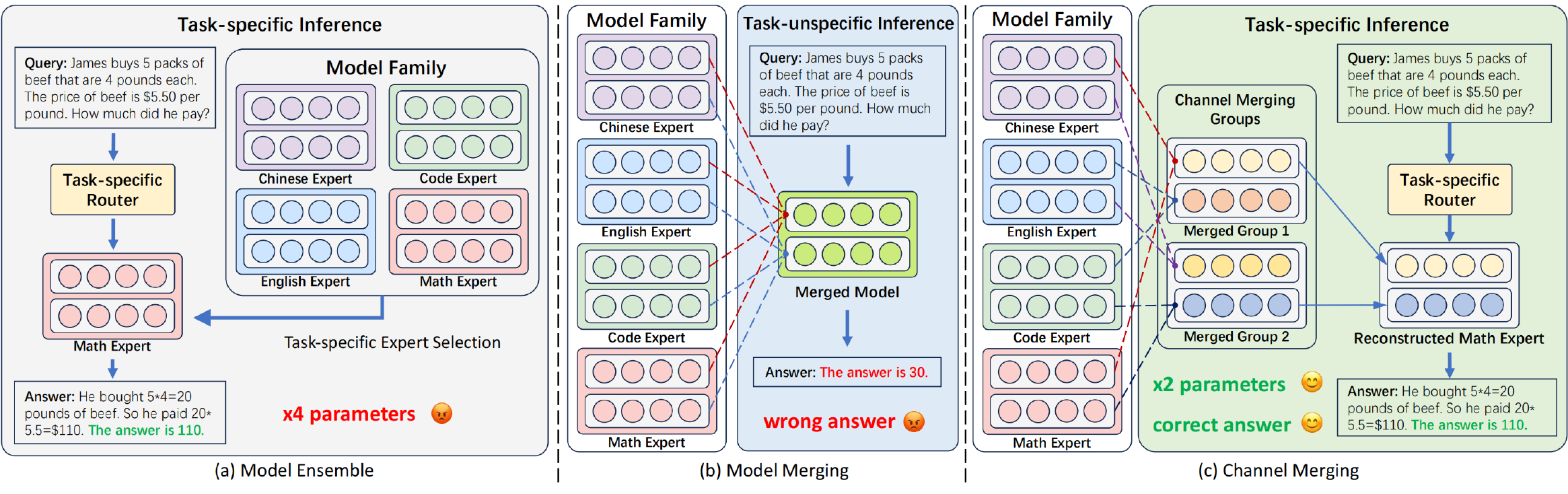}
    \vspace{-2em}
    \caption{ \label{fig:overview} This diagram contrasts various methods of handling multiple experts in LLMs. Panel (a) illustrates the conventional model ensemble approach, which requires loading all expert models into memory, leading to significant storage inefficiency. Panel (b) depicts the model merging strategy that simplifies the memory load but results in performance degradation due to parameter conflicts. Panel (c) presents our proposed Channel Merging method, which clusters and merges channel parameters, retaining each expert's unique features and ensuring efficient and effective performance.}
\end{figure*}

\section{Introduction}
Recent advancements in large language models (LLMs) such as LLaMA \cite{touvron2023llama} and Mistral \cite{jiang2023mistral} have significantly pushed the boundaries of artificial intelligence, achieving near-human performance across various general tasks. Despite these achievements, a notable performance gap remains in specialized domains such as coding and mathematics. Addressing this gap, many LLMs undergo task-specific fine-tuning to enhance their capabilities within these targeted areas \cite{Chinese-Mistral,yu2023metamath,wu2023pmc}. In multi-task 
scenarios,
it is common to ensemble LLMs specialized in different tasks to optimize performance. However, as shown in Figure \ref{fig:overview} (a), traditional ensembling methods \cite{tang2024merging,lu2023routing} require loading all specialized models into GPU memory, which is highly storage-intensive.

\begin{table}[htbp]
    \centering
    \scalebox{0.85}{
    \begin{tabular}{c|c|c}
    \toprule
    \toprule
         Method& Performance & Efficiency \\
        \midrule
         One-size-fit-all \cite{yu2023language}&$\times$&$\checkmark$\\ 
         Partial Merging \cite{jiang2023effective}&$\checkmark$&$\times$\\
         Channel Merging (Ours)&$\checkmark$&$\checkmark$\\
    \bottomrule
    \bottomrule
    \end{tabular}
    }
    \vspace{-0.5em}
    \caption{This table compares the scalability of different merging approaches in terms of maintaining performance and efficiency as the number of experts increases.}
    \label{table:comparison_pd_ed}
\end{table}

\noindent To combat the inefficiency of the model ensemble, model merging strategies \cite{yadav2024ties,yang2023adamerging} have been introduced, where delta parameters \cite{ilharco2022editing} from each expert are merged into the pre-trained parameters, allowing the system to load weights equivalent to a single expert during inference. Nevertheless, as shown in Figure \ref{fig:overview} (b), this one-size-fits-all merging can exacerbate parameter conflicts as the number of experts increases, often leading to a decline in downstream performance.
To mitigate parameter conflicts, previous approaches can be divided into two main strategies: (1) Post pruning, which prunes delta parameters—the alteration of the model parameters before and after fine-tuning \cite{yu2023language,yadav2024ties}. However, performance degradation still occurs as the number of merged experts increases. (2) Implementing partial merging \cite{jiang2023effective}, which merges task-agnostic parameters and maintains task-specific ones. 
Nonetheless, this strategy becomes less storage-efficient with the escalation in the number of experts, as it necessitates the retention of more separate parameters.

\noindent To effectively mitigate parameter conflicts while enhancing storage efficiency, we specifically analyze the layer-by-layer channel similarities between several experts and highlight that merging with finer granularity can further reduce parameter conflicts. Based on our analysis, we introduce a novel strategy called Channel Merging. As illustrated in Figure \ref{fig:overview}(c), this approach first clusters channel parameters that are across various experts into several groups based on their similarities, merging only those within identical groups to minimize conflicts. Subsequently, during inference, Channel Merging adaptively looks up the task-specific parameters required for each expert. Compared with model merging and model ensemble methods, Channel Merging mitigates performance degradation by retaining the unique knowledge of each expert model while reducing the total parameters loaded to GPU memory.
\begin{figure*}[t]
    \centering
    \includegraphics[scale=0.35]{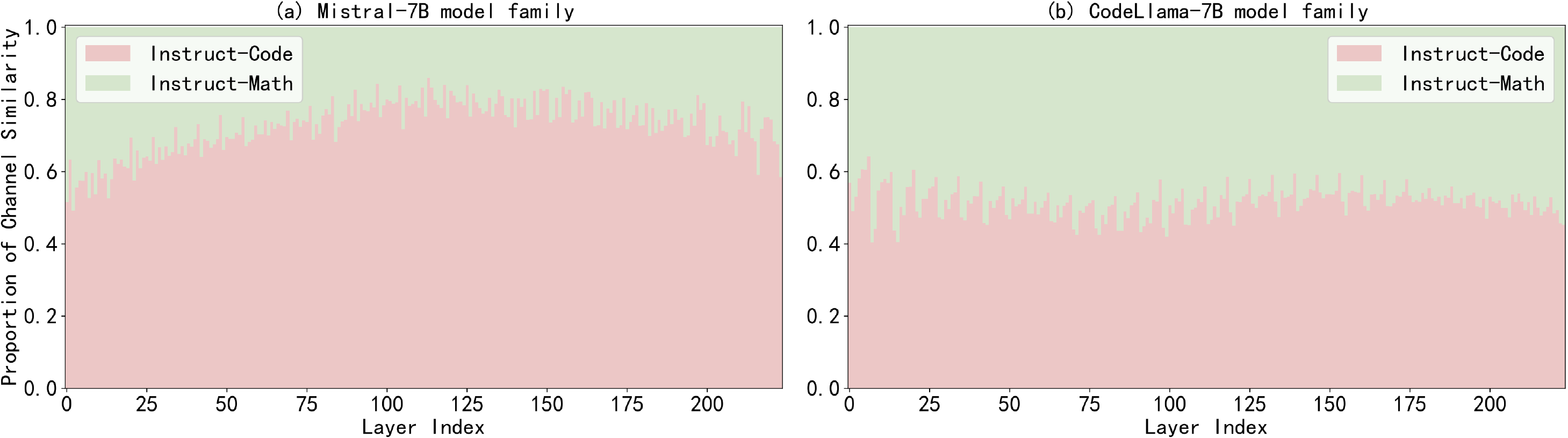}
    \caption{\label{fig:similarity} Layer-wise proportion of channel similarity between Instruction expert and other experts in (a) Mistral-7B model and (b) CodeLLaMA-7B families. The blue portions represent the proportion of channels in the Instruction expert that are more similar to the Code expert, while the red portions indicate the proportion of channels more similar to the Math expert.
    }
\end{figure*}

\noindent Our experiments, including specialized tasks like English reasoning, mathematic reasoning, code generation, and Chinese reasoning demonstrate that Channel Merging achieves performance on par with unmerged models. Additionally, we utilize a task-specific router to optimize expert selection for each query, enhancing the model's overall effectiveness. Our findings from general tasks show that Channel Merging, combined with the router, not only outperforms the Chinese-Mistral-7B-Instruct-v0.1 by 1.26\% on the AGIEval benchmark and achieves a 2.13\% improvement on the combined MMLU+CMMLU benchmarks but also requires only 53\% of the parameter load compared to traditional ensemble methods, clearly demonstrating its efficacy and versatility in various applications. Our contributions are mainly three-fold:
\begin{itemize}[leftmargin=*]
\item Through a detailed analysis of layer-by-layer channel similarities between different experts, we highlight the limitations of traditional model-level merging methods and demonstrate the necessity of finer-grained merging.

\item Based on our analysis, we introduce Channel Merging, a novel strategy that merges parameters at the channel level. This method mitigates parameter interference and maintains parameter efficiency as the number of experts increases.

\item The effectiveness of Channel Merging is demonstrated through extensive experimental results. For example, it shows minimal performance loss compared to one-size-fits-all and partial merging methods across various downstream tasks, such as English reasoning, mathematical reasoning, code generation, and Chinese reasoning. Additionally, when combined with a task-specific router, it achieves performance comparable to the model ensemble method on general tasks while requiring only 53\% of the parameters.
\end{itemize} 

\section{Related Work}
\textbf{Model merging.} 
As large pre-trained models are a repository of extensive knowledge, fine-tuning them for new tasks has become a prevalent method \cite{dodge2020fine}. Model merging, involving merging task-specific models fine-tuned from the same pre-trained model, has been increasingly recognized as an effective strategy to enhance generalization and multi-task capabilities in LLMs \cite{yadav2024ties,yang2023adamerging,yu2023language,matena2022merging,ilharco2022editing,ainsworth2022git,entezari2021role}. 
Although model merging provides enhanced flexibility and utility \cite{daheim2023model,NEURIPS2022_70c26937}, straightforward techniques like model averaging \cite{wortsman2022model} often lead to substantial performance degradation across multiple tasks due to parameter conflicts. To address these conflicts, strategies such as TIES Merging \cite{yadav2024ties} and DARE \cite{yu2023language} have been proposed, which involve pruning some delta parameters before the merging process. Nonetheless, parameter conflicts can still be stringent as the number of merged models increases. Another mitigation strategy is partial merging \cite{stoica2023zipit,kim2023solar,jiang2023effective}, which involves merging only a part of the parameters while preserving others independently. For example, ZipIT \cite{stoica2023zipit} selectively unmerges certain layers, effectively creating a multi-head model. Passthrough \cite{kim2023solar} concatenates layers from different LLMs, producing a deeper model. BYOM \cite{jiang2023effective} preserves some task-specific parameters according to magnitude. However, these partial merging methods become storage-inefficient as the number of merged experts increases. In contrast, our Channel Merging approach merges experts into fixed groups, thereby retaining storage efficiency even as the number of merged experts grows. To mitigate parameter conflicts, Channel Merging operates at the channel level and considers the similarity between channels of different experts. As shown in Table \ref{table:comparison_pd_ed}, Channel Merging preserves both performance and efficiency as the number of merged experts increases.

\noindent\textbf{Large language model ensemble.} 
LLM ensembling aims to combine off-the-shelf large language models (LLMs) to consistently improve performance across a variety of downstream tasks. LLM-BLENDER \cite{jiang2023llm} infers outputs from all candidate models and then ranks them using a reward function, which introduces significant computational overhead. To mitigate this, FrugalGPT \cite{chen2023frugalgpt} adopts a sequential inference process that stops as soon as it generates a response of sufficient quality, thereby reducing the need to infer from all models. Additionally, to cut down on computation further, several router-based methods \cite{tang2024merging,lu2023routing,shnitzer2023large} have been developed, which employs a trained routing function that accurately directs each query to the LLM best suited for it. Consequently, only one expert LLM is activated during each inference cycle. For instance, \citep{shnitzer2023large} shows the utility and limitations of learning model routers from various benchmark datasets. Zooter \citep{lu2023routing} distills a reward model to a task-specific router, assigning queries to experts more accurately. However, these methods often require preserving all expert models' weights in GPU memory, which can lead to memory inefficiencies. In contrast, our approach leverages the channel similarities between experts by merging multiple expert parameters into a few clusters, significantly reducing the parameter storage requirements during inference. Moreover, our method dynamically activates and reconstructs different experts based on the incoming query, maintaining expert diversity while minimizing the total parameters. 

\section{Preliminaries}
\textbf{Formulation of model merging.} 
Assuming a pretrained model, let $\boldsymbol{P} \in \mathbb{R}^{O\times I}$ represent the parameters of a specific layer, where $O$ and $I$ correspond to the output and input channel number, respectively.
From this model, we derive a set of $N$ task-specific models with parameters $\boldsymbol{\theta} = \{\boldsymbol{\theta}^{t_1}, \boldsymbol{\theta}^{t_2}, ..., \boldsymbol{\theta}^{t_N}\} \in \mathbb{R}^{N\times O\times I}$, each fine-tuned for a distinct task. Model merging is the process of integrating the modifications of all task-specific models back into a single model. This is achieved by first calculating the delta parameters for each task-specific model, which represent the changes made during the fine-tuning process relative to the pretrained model. These delta parameters are defined as $\boldsymbol{\delta}^{t_n} = \boldsymbol{\theta}^{t_n} - \boldsymbol{P}$ for each task $n$. Using the task arithmetic method \cite{ilharco2022editing}, the merged parameters, $\boldsymbol{\theta^{*}} \in \mathbb{R}^{O\times I}$, are computed as follows:
\begin{equation}
    \boldsymbol{\theta^{*}} = \boldsymbol{P} + \lambda \sum_{t_n} \boldsymbol{\delta}^{t_n},
\end{equation}
where $\lambda$ is a scaling factor that adjusts the influence of the delta parameters on the merged model. 

\noindent\textbf{Task-specific routing.}
To address computational efficiency concerns within LLM ensembles, task-specific routing is employed. This technique strategically selects a single LLM expert, denoted as $m^{*}$, that is best suited to respond to a given query $q$. This selection is determined through a scoring function that evaluates the appropriateness of each expert for the query, formalized as follows:
\begin{equation}
    m^{*} = \underset{m\in M}{\mathrm{argmax}} \ \mathcal{Z}(q, m),
\end{equation}
where $M$ represents the set of all available LLM experts, and $\mathcal{Z}(\cdot)$ is a function that scores each expert $m$ based on its predicted effectiveness in responding to query $q$. The expert with the highest score is chosen for the task, optimizing the ensemble’s computational resources by activating only the most relevant model.

\begin{figure*}[htbp]
    \centering
    \includegraphics[width=1\linewidth]{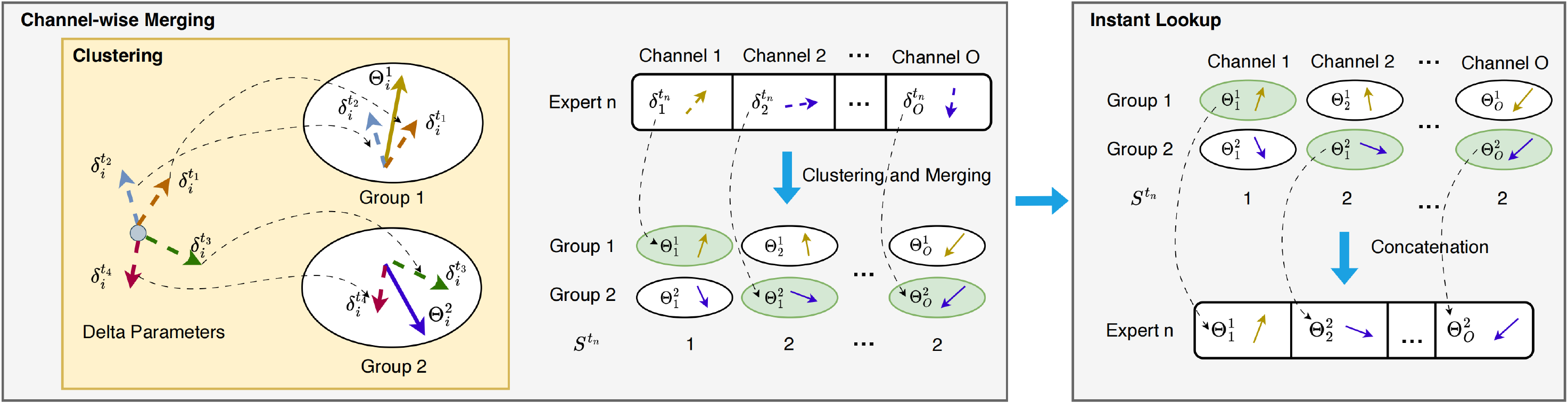}
    \caption{\label{fig:method}An illustration of our Channel Merging method. The process involves two core parts: Channel-wise Merging and Instant Lookup. In the channel-wise Merging stage, for each channel, delta parameters from each expert $\mathbf{\delta}_{i}^{t_n}$ are clustered into different groups, and parameters within the same group are merged into $\mathbf{\Theta}_{i}^{k}$. $S^{t_n}$ records the group index that each expert’s parameters have been merged into (shown in ForestGreen). During inference, the activated expert can instantly look up its parameters from the groups based on $S^{t_n}$.}
\end{figure*}

\section{Method}
In this section, we begin by analyzing the inconsistency in channel similarity across different expert models, illustrating how simplistic merging strategies—such as merging models at the model level—can be sub-optimal. Subsequently, we introduce Channel Merging, a channel-wise merging strategy tailored to optimize both parameter compression and model performance. As illustrated in Figure \ref{fig:overview} (c), Channel Merging groups parameters from different experts in the model family according to their channel similarities. During the inference, the parameters of the activated expert are instantly looked up from these offline merged groups, ensuring efficient and effective performance.

\subsection{Similarity Inconsistency}
Before we delve into Channel Merging, it’s essential to address a fundamental question: Why merge at the channel level instead of the model level, as was common in previous methods? To answer this, we specifically analyze the layer-by-layer channel similarities between the Instruction expert and the other two experts—Code and Math. This analysis involves quantifying how many channels in the Instruction expert are more similar to the Code expert compared to the Math expert based on their cosine similarities. We employ cosine similarity to measure the similarity between parameters since high cosine similarity between neural network parameters correlates with similar activations produced by the layers \cite{mason2024makes,klabunde2023towards}. The details of LLM candidates used for comparison can be found in \textbf{Appendix}.

\noindent In the Mistral-7B model family (Figure \ref{fig:similarity}(a)), overall, the Instruction expert channels show closer similarity to the Code expert channels, with the proportion of more similar channels typically being higher. However, even in the layers where the similarity proportion peaks, more than 20\% of Instruction channels are closer to the Math expert, highlighting significant similarity inconsistency between those experts. Moreover, the CodeLLaMA-7B model family (Figure \ref{fig:similarity}(b)) exhibits more random variations in similarity. Approximately half of the channels in all layers show greater similarity to the Code expert, while the other half are more aligned with the Math expert. These experiments highlight a crucial insight: \textbf{Finer granularity, such as channel-level merging, can further reduce parameter conflicts arising from insufficiently similar parameters.}

\subsection{Merging with Channel Similarity}
To effectively merge multiple expert LLMs while accommodating significant variations in channel similarities, we introduce a novel method termed Channel Merging. As illustrated in Figure \ref{fig:method}, our method unfolds in two parts: Channel-wise Merging and Instant Lookup.

\begin{table*}[htbp]
\renewcommand{\arraystretch}{1.3}
\caption{Comparison on downstream tasks. `Baseline' refers to the performance metrics of experts when unmerged. TIES-CM and DARE-CM represent TIES and DARE methods combined with Channel Merging, respectively.}
\vspace{-0.5em}
\centering
\scalebox{0.75}
{
\begin{tabular}{c|ccc|ccc|ccc|cccc}
\toprule
\toprule
\multirow{2}{*}{Method} & \multicolumn{3}{c}{Instruction Expert (\%) $\uparrow$} & \multicolumn{3}{c}{Math Expert (\%) $\uparrow$} & \multicolumn{3}{c}{Code Expert (\%) $\uparrow$} & \multicolumn{3}{c}{Chinese Expert (\%) $\uparrow$} \\
\cmidrule(l){2-4} \cmidrule(l){5-7} \cmidrule(l){8-10} \cmidrule(l){11-13}
& CommonSenseQA & TriviaQA & Avg. & GSM8K & Math & Avg. & HumanEval & MBPP & Avg.& CMMLU & CEVAL & Avg. \\
\midrule
Baseline& 75.86& 58.39 & 67.12&73.92&20.62&47.27&45.36&43.20&44.28&47.52&47.50&47.51 \\
\midrule
BYOM&75.67&63.86&69.77&65.08&16.95&41.02&43.62&40.23&41.93&48.15&47.83&48.00\\
TIES &70.62&50.06&60.34&63.51&10.84&37.17&30.26&35.80&33.03&43.05&45.80&44.42\\
DARE&73.85&51.93&62.89&63.02&10.38&36.70&31.58&36.26&33.92&44.32&45.96&45.14\\
TIES-CM (Ours) &73.61&60.85&67.23&68.14&18.26&43.2&43.28&41.81&42.55&47.93&47.38&47.66\\
DARE-CM (Ours)&\textbf{75.27}&\textbf{64.49}&\textbf{69.88}&\textbf{70.05}&\textbf{19.89}&\textbf{44.95}&\textbf{45.13}&\textbf{43.00}&\textbf{44.06}&\textbf{48.41}&\textbf{47.69}&\textbf{48.05}\\
DARE-CM + router (Ours)&75.27&64.49&69.88&70.01&19.85&44.93&45.12&43.00&44.06&48.41&47.69&48.05\\
\bottomrule
\bottomrule
\end{tabular}
}
\label{table:results_downstream_tasks}
\end{table*}

\noindent\textbf{Channel-wise merging.} In this stage, we cluster the delta parameters for each output channel from different experts based on cosine similarity. Specifically, for the $i$-th output channel, delta parameters across all experts $\boldsymbol{\delta}_{i} = \{\boldsymbol{\delta}_{i}^{t_1}, \boldsymbol{\delta}_{i}^{t_2}, ..., \boldsymbol{\delta}_{i}^{t_N}\} \in \mathbb{R}^{N \times I}$ are grouped using the K-Means clustering algorithm into $K$ clusters $\boldsymbol{C}_{i} = \{\boldsymbol{C}_{i}^{1}, \boldsymbol{C}_{i}^{2}, ..., \boldsymbol{C}_{i}^{K}\}$ ($K < N$).
This clustering ensures that each group contains parameters with high similarity, thus reducing conflicts during merging. These similar parameters within each cluster $k$ are then merged with the task arithmetic \cite{ilharco2022editing}:
\begin{equation}
    \boldsymbol{\Theta}_{i}^k = \boldsymbol{P}_{i} + \lambda \sum_{\boldsymbol{\delta}_{i}^{t_n} \in \boldsymbol{C}_{i}^k} {\boldsymbol{\delta}_{i}^{t_n}}.
    \label{eq:merge}
\end{equation}
Repeating the clustering and merging for all channels in the layer, we can obtain $K$ groups new parameters $\mathbf{\Theta} \in \mathbb{R}^{K\times O\times I}$, where the $k$-th group parameter $\boldsymbol{\Theta}^k$ can be represented as:
\begin{equation}
    \boldsymbol{\Theta}^k \in \mathbb{R}^{O\times I} = \{\boldsymbol{\Theta}_{1}^k, \boldsymbol{\Theta}_{2}^k,...,\boldsymbol{\Theta}_{O}^k\}.
\end{equation}
Additionally, an index set $S^{t_n}=\{S_{1}^{t_n}, ...,S_{O}^{t_n}\} \in \mathbb{R}^{1\times O}$, where $S_{i}^{t_n} \in \{1, ..., K\}$, is maintained for each expert $t_n$, indicating which group their channel parameters are merged into. It is worth noting that the Channel-wise Merging stage is executed offline, thus imposing no additional computational overhead during inference.

\noindent\textbf{Instant lookup.} Channel Merging merges different channels from each expert into distinct groups, thereby preserving the unique characteristics of each expert compared to one-size-fits-all merging. During inference, we can selectively activate the expert that is the most suitable to respond to the input query. Since the merged parameter groups and expert index are constructed offline in the Channel-wise Merging stage, the parameter of the activated expert can be instantly looked up and concatenated from the corresponding groups according to the index stored in $S^{t_n}$. Specifically, the layer-level parameters of activated experts $\boldsymbol{\hat{\theta}}^{t_n} \in \mathbb{R}^{O\times I}$ can be formally written as:
\begin{equation}
    \boldsymbol{\hat{\theta}}^{t_n} = \underset{S_{i}^{t_n} \in S^{t_n}}{\bigoplus} \boldsymbol{\Theta}_{i}^{S_{i}^{t_n}},
\end{equation}
where $\bigoplus$ denotes the concatenation operation, which aligns the channel parameters from selected groups to reconstruct the full parameter set for each layer of the $t_n$-th expert. This concatenation ensures that the structural and functional characteristics of each expert's layers are maintained, while only the parameters of the relevant expert are activated.

\noindent\textbf{Model size reduction analysis.} Assuming each expert in a model zoo has $\Psi$ parameters, managing $N$ distinct experts would require storing $N\Psi$ parameters separately. In contrast, our Channel Merging method organizes each expert’s parameters by clustering them along the output channel dimension into $K$ distinct categories, each containing a full set of $\Psi$ parameters. Consequently, the total storage required for parameters is reduced to $K\Psi$. Although each expert maintains an index $S^{t_n}$, the size of this index is considerably smaller than $\Psi$—equivalent to the total number of channels—therefore, it can be considered negligible in the overall parameter count. With the implementation of Channel Merging, the total number of parameters necessary is effectively diminished to $\frac{K\Psi}{N\Psi} = \frac{K}{N}$ ($K < N$). Since $K$ is a constant, unlike partial merging, Channel Merging can maintain storage efficiency even as the number of merged experts increases.
\subsection{Task-specific Routing}
As shown in Figure \ref{fig:overview}(c), following the paradigm of previous LLM ensemble methods \cite{liu2024meswitch}, we employ a task-specific router to determine which expert to activate and reconstruct for a given query. To train this router efficiently, we operate under the assumption that an expert will perform optimally on queries that originate from its fine-tuning dataset. To implement this, we sample a set of queries $Q$ from the datasets of various tasks, using the originating task classes (e.g., code, math, instruction, Chinese) as the label $Y$. The optimization process for training the router is then defined as follows:
\begin{equation}
    \mathcal{Z}^* = \underset{\mathcal{Z}}{\mathrm{argmin}} \sum_{(q, y)\in (Q, Y)} -y\cdot log(\mathcal{Z}(q, m))
\end{equation}
In this equation, $\mathcal{Z}(q,m)$ calculates the probability that expert $m$ is the most suitable for handling query $q$, based on the learned task-specific affinities. Once this model is trained, the arrival of a new query triggers the task-specific router, which employs the optimized function $\mathcal{Z}^{*}$ to determine which expert should be activated. This process ensures that each query is handled by the expert most likely to achieve the best performance, thereby enhancing overall efficiency and effectiveness.

\section{Experiments}
\subsection{Experimental Setting}
\noindent\textbf{Test datasets.}
To evaluate the performance of merging,
we report accuracy on several benchmarks across different domains: CommonSenseQA \cite{talmor2018commonsenseqa} and TriviaQA \cite{joshi2017triviaqa} for the instruction, GSM8K \cite{cobbe2021training} and MATH \cite{hendrycks2021measuring} for the mathematics, HumanEval \cite{chen2021evaluating} and MBPP \cite{austin2021program} for the code, and CEval \cite{huang2024c} and CMMLU \cite{li2023cmmlu} for the Chinese. Besides, we evaluate the merged model with the task-specific router on several general task benchmarks: MMLU \cite{hendrycks2020measuring}, CMMLU, and AGIEval \cite{zhong2023agieval}. We use the OpenCompass toolbox \cite{contributors2023opencompass} to evaluate all datasets.

\begin{table*}[htbp]
\renewcommand{\arraystretch}{1.3}
\caption{Comparison of general tasks across multiple domains. "Total Param." and "Activate Param." refer to the total number and the activated number of parameters, respectively.}
\vspace{-0.5em}
\centering
\scalebox{0.75}
{
\begin{tabular}{ccc|cccc|cccccc}
\toprule
\toprule
\multirow{2}{*}{Model}& \multirow{2}{*}{Total} & \multirow{2}{*}{Activate} &\multicolumn{4}{c}{AGIEval (\%) $\uparrow$} & \multicolumn{5}{c}{MMLU+CMMLU (\%) $\uparrow$} \\
\cmidrule(l){4-7} \cmidrule(l){8-13}
&Param. (B)& Param. (B)& Chinese & English & Gaokao & Avg. & Humanities & Social & Stem&Other & Chinese & Avg. \\
\midrule
Dolphin-2.2.1-Mistral-7B& 6.7 & 6.7&32.30&39.21&34.84&35.45&56.26&59.61&44.63&57.76&41.80&52.01\\
MetaMath-Mistral-7B&6.7&6.7&31.65&37.63&34.53&34.60&54.91&	58.615&	43.675&	57.35&	40.3&	50.97\\
Speechless-Code-Mistral-7b-V1.0&6.7&6.7&32.53&39.16&35.89&35.86&55.67&	59.67&	44.36&	57.89&	40.30&	51.58\\
Chinese-Mistral-7B-Instruct-v0.1&6.7&6.7&36.10&37.09&37.30&36.75&56.30&	49.95&	51.65&	56.75&	\textbf{46.80}&	52.29\\
\midrule
BYOM + router &14.7&6.7&35.51&37.93&39.41&37.62&57.86&61.83&50.03&55.97&46.08&54.35\\
Model Ensemble + router &26.8&6.7&\textbf{36.10}&\textbf{39.21}&39.44&\textbf{38.25}&\textbf{58.30}&60.32&\textbf{51.72}&57.04&44.80&\textbf{54.43}\\
DARE-CM + router (Ours) &\textbf{14.3}&6.7&35.99&38.49&\textbf{39.57}&38.01&58.29&\textbf{62.51}&46.14&\textbf{59.49}&45.69&54.42\\
\bottomrule
\bottomrule
\end{tabular}
}
\label{table:results_general_tasks}
\end{table*}
\begin{table*}[h]
\renewcommand{\arraystretch}{1.3}
\caption{Experimental results on different merging granularities.}
\centering
\scalebox{0.77}
{
\begin{tabular}{c|ccc|ccc|ccc|cccc}
\toprule
\toprule
\multirow{2}{*}{Granularity} & \multicolumn{3}{c}{Instruction Expert (\%) $\uparrow$} & \multicolumn{3}{c}{Math Expert (\%) $\uparrow$} & \multicolumn{3}{c}{Code Expert (\%) $\uparrow$} & \multicolumn{3}{c}{Chinese Expert (\%) $\uparrow$} \\
\cmidrule(l){2-4} \cmidrule(l){5-7} \cmidrule(l){8-10} \cmidrule(l){11-13}
& CommonSenseQA & TriviaQA & Avg. & GSM8K & Math & Avg. & HumanEval & MBPP & Avg.& CMMLU & CEVAL & Avg. \\
\midrule
Channel&75.27&\textbf{64.49}&\textbf{69.88}&\textbf{70.05}&\textbf{19.89}&\textbf{44.95}&\textbf{45.13}&\textbf{43.00}&\textbf{44.01}&48.41&\textbf{47.69}&\textbf{48.05} \\
Layer&74.85&63.80&69.33&67.70&18.26&42.98&43.28&41.33&42.31&\textbf{48.45}&47.53&47.99 \\
Model&\textbf{75.61}&61.20&68.41&66.20&15.20&40.7&40.25&40.20&40.23&47.38&47.20&47.29\\
\bottomrule
\bottomrule
\end{tabular}
}

\label{table:ablation_granularities}
\end{table*}

\noindent\textbf{Implementation details.} For model merging, we cluster the expert weights into several groups. Subsequently, we use the commonly used model merging algorithms from MergeKit \cite{goddard2024arcee} to merge the parameters in the same group: (1) \textbf{DARE-CM}, we randomly prune 30\% of the delta parameters for each expert before merging. (2) \textbf{TIES-CM},  we prune 30\% of the delta parameters based on their magnitude and sign for each expert before merging. $\lambda$ in Eq. (\ref{eq:merge}) is set to 0.5.  Unless otherwise specified, we define the number of groups as two. The detail of model candidates can be found in \textbf{Appendix}. The merging experiments can be done on only a single A100 GPU. To fairly assess the performance loss due to model merging, we only activate the corresponding expert for each task during downstream task testing. For general tasks, we use a task-specific router to activate different experts based on the task requirements. We show the detail of the task-specific router in \textbf{Appendix}.

\noindent\textbf{Contenders.} 
For the one-size-fit-all merging strategy, we compare with DARE \cite{yu2023language} and TIES \cite{yadav2024ties} which use the same post-pruning strategy as we mentioned in \textbf{implementation details} and merge all experts into one model. For the partial merging strategy, we compare our method with BYOM \cite{jiang2023effective}, which retains the top 30\% of parameters by magnitude for each expert and merges the remaining parameters into one group.

\begin{table*}[htbp]
\renewcommand{\arraystretch}{1.3}
\caption{Experimental results on different cluster methods.}
\centering
\scalebox{0.77}
{
\begin{tabular}{c|ccc|ccc|ccc|cccc}
\toprule
\toprule
\multirow{2}{*}{Cluster Method} & \multicolumn{3}{c}{Instruction Expert (\%) $\uparrow$} & \multicolumn{3}{c}{Math Expert (\%) $\uparrow$} & \multicolumn{3}{c}{Code Expert (\%) $\uparrow$} & \multicolumn{3}{c}{Chinese Expert (\%) $\uparrow$} \\
\cmidrule(l){2-4} \cmidrule(l){5-7} \cmidrule(l){8-10} \cmidrule(l){11-13}
& CommonSenseQA & TriviaQA & Avg. & GSM8K & Math & Avg. & HumanEval & MBPP & Avg.& CMMLU & CEVAL & Avg. \\
\midrule
KMeans&75.27&64.49&69.88&\textbf{70.05}&\textbf{19.89}&\textbf{44.95}&\textbf{45.13}&\textbf{43.00}&\textbf{44.01}&\textbf{48.41}&47.69&\textbf{48.05} \\
Random&75.01&63.45&69.23&68.84&19.07&43.96&44.21&41.00&42.61&47.82&46.05&46.94 \\
Sign&\textbf{75.93}&\textbf{64.96}&\textbf{70.45}&68.69&19.73&44.21&45.01&42.1&43.56&48.09&\textbf{47.93}&48.01\\
\bottomrule
\bottomrule
\end{tabular}
}
\label{table:ablation_cluster}
\end{table*}
\begin{figure*}[h]
    \centering
    \includegraphics[scale=0.35]{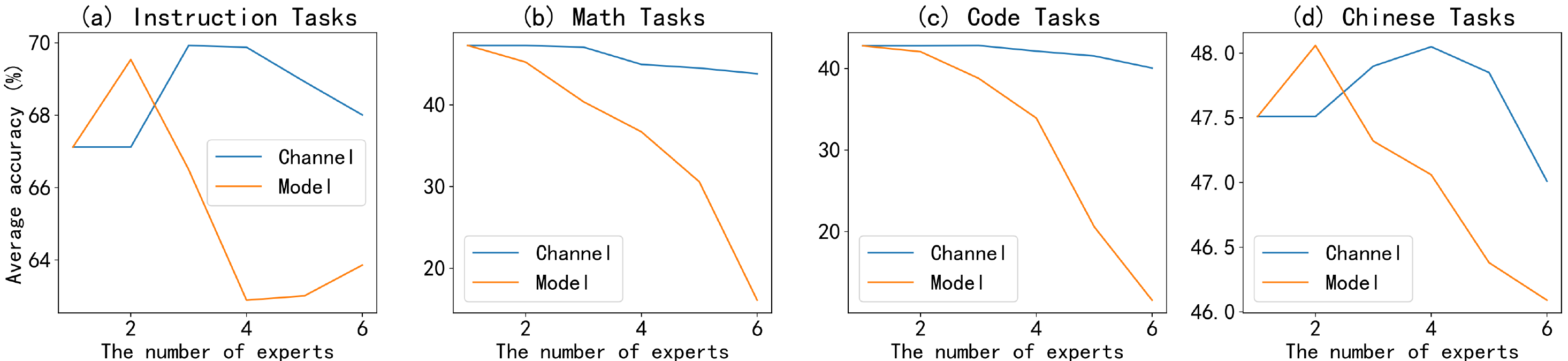}
\caption{\label{fig:ablation_num_experts} Experimental results on four different task categories as the number of experts varies from one to six. 'Channel' and 'Model' represent the accuracy achieved with channel-level and model-level merging, respectively.}
\vspace{-0.5em}
\end{figure*}
\subsection{Main Results}
\noindent\textbf{Downstream tasks.} 
To assess the necessity of channel-level merging, we conduct experiments across various downstream tasks to compare different merging methods, with and without the incorporation of channel merging. We use unmerged-downstream models as our baseline to establish a clear performance benchmark. The results, as shown in Table \ref{table:results_downstream_tasks}, highlight the effectiveness of channel-level merging. For example, in the Instruction Expert tasks, DARE-CM improves performance significantly, achieving an average score of 69.88\%, compared to 62.89\% with the model-level DARE. This represents a substantial 6.99\% increase in performance, underscoring the reduced performance degradation offered by our channel-level approach. Similarly, in the Code Expert tasks, DARE-CM scored 44.01\% on average, outperforming DARE by 7.31\%. When comparing DARE-CM to BYOM, we observe that DARE-CM consistently outperforms BYOM across various metrics. For instance, in the Math Expert tasks, DARE-CM achieves an average score of 44.95\%, compared to BYOM's 41.02\%. Notably, our DARE-CM yields the most optimal results, maintaining performance levels close to or even surpassing the baseline in specialized tasks such as Instruction and Chinese tasks. These findings demonstrate that our method not only mitigates the performance loss but can also enhance the model’s effectiveness in handling domain-specific queries. We also compare the performance metrics with and without the use of the trained router on DARE-CM. The results in Table \ref{table:results_downstream_tasks} clearly demonstrate that the router's deployment maintains the effectiveness of our model.  

\noindent\textbf{General tasks.} Given that the goal of model merging is to achieve a more versatile model, we integrate our merged models DARE-CM with a task-specific router and compare their performance on general tasks against the unmerged baseline models. As indicated in Table \ref{table:results_general_tasks}, these benchmarks encompass a variety of tasks including those in Chinese, English, mathematics, and coding. Consequently, Channel Merging combined with the router consistently outperforms the separate unmerged models. For example, Channel Merging exhibited a 1.26\% higher average accuracy over the Chinese-Mistral-7B-Instruct-v0.1 in the AGIEval benchmark, and a 2.13\% improvement in the combined MMLU+CMMLU benchmarks. Moreover, we find that DARE-CM can surpass BYOM in several benchmarks. For instance, in the AGIEval and MMLU+CMMLU benchmarks, DARE-CM + router achieves an average score of 38.01\% and 54.42\%, notably higher than the 37.62\% and 37.62\% of BYOM + router, respectively. When compared to the traditional model ensemble with the router, Channel Merging achieves comparable outcomes across all benchmarks, requiring only 53\% of the parameter used by traditional methods.

\subsection{Ablation Studies}
\noindent\textbf{Merging granularities.} A key distinction of our Channel Merging approach compared to other merging methods lies in the granularity at which the merge is executed—specifically, at the channel level. To assess the impact of different merging granularities on model performance, we conducted an ablation study. This study involved merging operations at three different levels: Channel, Layer, and the entire Model, with subsequent performance evaluation on downstream tasks. As shown in Table \ref{table:ablation_granularities}, channel-level merging outperformed both layer-level and model-level merging across various tasks, achieving the highest average performance metrics. This suggests that finer granularity in merging helps to reduce parameter conflicts, thereby preserving more of the downstream task performance.

\noindent\textbf{The sensitivity to the merged experts number.} To explore how the number of experts influences the effectiveness of Channel Merging, we carry out experiments that assessed the performance across various downstream tasks when integrating varying numbers of expert models. We expand our set of candidate models to include Hercules-2.5-Mistral-7B and CollectiveCognition-v1.1-Mistral-7B, allowing for the integration of up to six experts. Additionally, we compare channel-level merging with results from model-level merging.
The experimental results, as depicted in the figures, show that, in contrast to model-level merging, channel-level merging exhibits a markedly slower rate of performance degradation as the number of merged experts increases, particularly in tasks involving mathematic reasoning and code generation. 

\noindent\textbf{The effect of clustering methods.} To validate the appropriateness of using the KMeans method for clustering channel parameters, we compare its impact on experimental results with two alternative clustering strategies: (1) Random, where channels are grouped randomly, and (2) Sign \cite{yadav2024ties}, where parameters are grouped based on having the same sign. The results, as shown in Table \ref{table:ablation_cluster}, reveal that both KMeans and Sign clustering significantly outperform the random grouping method. This indicates that logically group parameters (either by minimizing intra-cluster variance in KMeans or aligning parameter signs) lead to better performance than arbitrary grouping.

\section{Conclusion}
In this paper, we introduced Channel Merging, a novel strategy designed to enhance the efficiency and performance of merging LLMs specialized in various tasks. By clustering and merging channel parameters based on their similarities, Channel Merging mitigates the parameter conflicts associated with traditional one-size-fits-all merging methods. Through extensive experiments, we have demonstrated that Channel Merging achieves comparable performance to unmerged experts in tasks such as English reasoning, mathematical reasoning, code generation, and Chinese reasoning. Additionally, when integrated with a task-specific router, Channel Merging outperforms traditional ensemble methods in general tasks while requiring only 53\% of the parameters, showcasing significant improvements in both performance and storage efficiency.

\noindent\textbf{Limitation and future work.}
Channel Merging requires that the experts to be merged are fine-tuned from the same pretrained model. Additionally, compared to one-size-fits-all approaches, Channel Merging may increase the parameters of the merged model. In future work, we plan to explore further compression of the merged model's parameter size by setting different groups for each layer.

\bibliography{aaai25}

\newpage

\appendix

\label{sec:appendix}

\begin{center}
	{
		\Large{\textbf{Appendix}}
	}
\end{center}
\begin{table*}[!htbp]
\renewcommand{\arraystretch}{1.3}
\caption{\label{table:merging_groups}Experimental results on different merging groups.}
\centering
\scalebox{0.65}
{
\begin{tabular}{c|c|ccc|ccc|ccc|cccc}
\toprule
\toprule
\multirow{2}{*}{Groups} & \multirow{2}{*}{Param. (B)} &\multicolumn{3}{c}{Instruction Expert (\%) $\uparrow$} & \multicolumn{3}{c}{Math Expert (\%) $\uparrow$} & \multicolumn{3}{c}{Code Expert (\%) $\uparrow$} & \multicolumn{3}{c}{Chinese Expert (\%) $\uparrow$} \\
\cmidrule(l){3-5} \cmidrule(l){6-7} \cmidrule(l){9-11} \cmidrule(l){12-14}
& &CommonSenseQA & TriviaQA & Avg. & GSM8K & Math & Avg. & HumanEval & MBPP & Avg.& CMMLU & CEVAL & Avg. \\
\midrule
1&6.7&73.85&51.93&62.89&63.02&10.38&36.70&31.58&36.26&33.92&44.32&45.96&45.14\\
2&14.3&75.27&\textbf{64.49}&\textbf{69.88}&70.05&19.89&44.95&45.13&43.00&44.01&\textbf{48.41}&\textbf{47.69}&\textbf{48.05} \\
3&21.9&75.36& 60.21 & 67.79&71.83&20.43&46.13&45.10&43.46&44.28&47.52&47.36&47.44 \\
4&26.8& \textbf{75.86}& 58.39 & 67.12&\textbf{73.92}&\textbf{20.62}&\textbf{47.27}&\textbf{45.36}&\textbf{43.20}&\textbf{44.28}&47.52&47.50&47.51 \\

\bottomrule
\bottomrule
\end{tabular}
}
\end{table*}
\begin{table*}
\centering
\caption{\label{table:llm_detail}Versions and correspondences with pre-trained backbones of fine-tuned LLMs.}
\scalebox{0.85}
{
\begin{tabular}{@{}lll@{}}
\toprule
Tasks & Fine-tuned LLMs & Pre-Trained backbones \\ \midrule
English Reasoning &  Dolphin-2.2.1-Mistral-7B \cite{dolphin-2.2.1-mistral-7b} & Mistral-7B-v0.1 \cite{jiang2023mistral} \\
                       &  Hercules-2.5-Mistral-7B \cite{Hercules-2.5-Mistral-7B} & Mistral-7B-v0.1 \cite{jiang2023mistral} \\
                       &  CollectiveCognition-v1.1-Mistral-7B \cite{CollectiveCognition-v1.1-Mistral-7B} & Mistral-7B-v0.1 \cite{jiang2023mistral} \\
                       &CodeLlama-7b-Instruct \cite{roziere2023code}& CodeLlama-7b \cite{roziere2023code} \\ \midrule
Mathematical Reasoning &  MetaMath-Mistral-7B \cite{MetaMath-Mistral-7B} & Mistral-7B-v0.1 \cite{jiang2023mistral} \\
                       & 
OpenMath-CodeLlama-7b-Python \cite{toshniwal2024openmath} & CodeLlama-7b \cite{roziere2023code}\\ \midrule
Code Generation        & Speechless-Code-Mistral-7B \cite{speechless-code-mistral-7b-v1.0}  & Mistral-7B-v0.1 \cite{jiang2023mistral} \\
                       & WizardCoder-Python-7B \cite{luo2023wizardcoder}& CodeLlama-7b \cite{roziere2023code}\\ \midrule
Chinese Reasoning      &  Chinese-Mistral-7B-Instruct-v0.1 \cite{Chinese-Mistral-7B-Instruct-v0.1}& Mistral-7B-v0.1 \cite{jiang2023mistral} \\
        
\bottomrule
\end{tabular}
}
\end{table*}
\section{Detailed Experimental Setting}
\subsection{Candidate LLMs}
\label{sec:candidate_llm}
To analyze the similarity relationship between different experts, we conduct experiments under Mistral-7B-v0.1 model family and CodeLLaMA model family. In Mistral-7B-v0.1 model family experiments, we use Dolphin-2.2.1-Mistral-7B as the instruction expert, MetaMath-Mistral-7B as the math expert and Speechless-Code-Mistral-7B as the code expert. In CodeLLaMA model family experiments, we use CodeLlama-7b-Instruct as the instruction expert, OpenMath-CodeLlama-7b-Python as the math expert and WizardCoder-Python-7B as the code expert.

\noindent In our main experiments, we apply our method to the Mistral-7B-v0.1 model family, including several specialized LLMs: Dolphin-2.2.1-Mistral-7B as the instruction expert, Speechless-Code-Mistral-7B as the code expert, MetaMath-Mistral-7B as the math expert, and Chinese-Mistral-7B-Instruct-v0.1 as the Chinese expert. These models are all fine-tuned derivatives of the foundational pretrained model Mistral-7B-v0.1. The details of each expert can be found in Table \ref{table:llm_detail}.

\subsection{Training detail of task-specific router} 
\label{sec:training_detail}
To efficiently assign queries to the optimal expert, We utilize the bert-base-multilingual-cased \cite{bert-base-multilingual-cased} which is a tiny model with only 110M parameters as the task-specific router. To train the task-specific router, we collect 50k instruction samples from various open-source datasets and randomly select 50k samples to train the router, including Dolphin \cite{Dolphin} for the instruction domain, MetaMathQA \cite{MetaMath-Mistral-7B} for the mathematics domain, WizardLM-evol-instruct-V2-196k \cite{WizardLM-evol-instruct-V2-196k} for the code domain, and Wizard-LM-Chinese-instruct-evol \cite{Wizard-LM-Chinese-instruct-evol} for the Chinese domain. We use the AdamW optimizer with $\beta_1 = 0.9$ and $\beta_2=0.95$ to train 1 epoch on a single A100GPU, setting the learning rate to $1\times10^{-4}$, the batchsize to 128, and applying a linear learning rate warmup. 

\begin{figure}[htbp]
    \centering
    \includegraphics[scale=0.5]{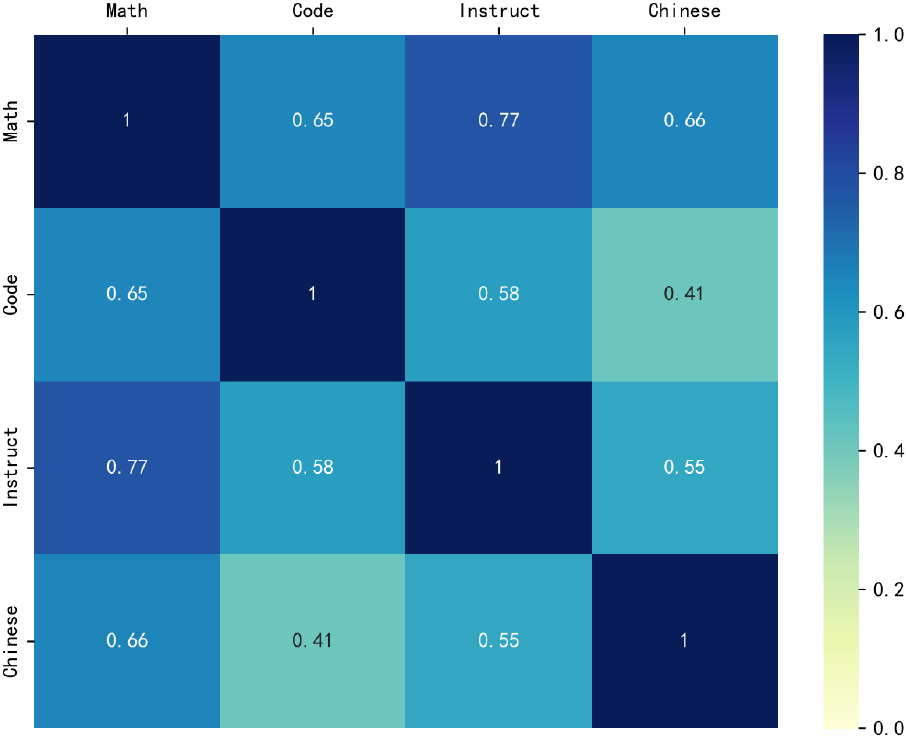}
    \caption{\label{fig:experts_similarity} Heatmap of channel similarities between different expert models. }
\end{figure}

\section{More Ablation Studies}
\noindent\textbf{The similarity between merged experts.} We assess the channel similarities between different experts after merging. The similarity between experts is calculated by counting the number of channels where $S^{t_n}$ values matched between two experts and then normalizing this count by the total number of channels. The results, depicted in Figure \ref{fig:experts_similarity}, show relatively low similarity scores across different experts. For instance, the Instruction and Math experts, despite having the highest similarity score in the matrix, still exhibit a moderate similarity of 0.77, suggesting that while there are shared characteristics, distinct features prevail. On the other end, the Code and Chinese experts have the lowest similarity of 0.41, indicating significant differences in their channel characteristics. These findings validate Channel Merging's ability to preserve expert uniqueness, effectively minimizing parameter conflicts and enhancing performance by ensuring each expert's specialized knowledge remains intact within the merged framework.

\noindent\textbf{Merging under different groups.} In Channel Merging, we can categorize expert parameters into varying numbers of groups. We conduct experiments to explore the impact of the number of groups on the merged model's parameter count and performance. The results, as shown in Table \ref{table:merging_groups}, indicate that when the number of groups is set to one, Channel Merging functions similarly to DARE \cite{jiang2023llm}. Although this setting minimizes the number of parameters required, it also yields the poorest performance. Conversely, when the number of groups is set to four, Channel Merging resembles a model ensemble, requiring the maximum number of parameters. Notably, setting the number of groups to two not only surpasses or closely matches the performance of having four groups across various downstream tasks but also significantly reduces the number of parameters needed.

\noindent\textbf{Latency analysis for Instant Lookup.} To comprehensively analyze the cost associated with the lookup and concatenation processes during inference, we randomly selected 50 data points from the MMLU validation dataset to evaluate the time costs associated with the lookup and inference processes. The final reported times for lookup and inference are the average values derived from these 50 data points. The results given in Table \ref{tab:inference_time} clearly show that the lookup and concatenation process accounts for a minimal portion of the total inference time.

\begin{table}[]
    \centering
    \begin{tabular}{c|c}
    \toprule
         Lookup and concatenation&Total inference  \\
    \hline
         0.06s& 0.831s\\
    \bottomrule
    \end{tabular}
    \caption{The inference time of the lookup and concatenation process.}
    \label{tab:inference_time}
\end{table}

\noindent\textbf{The effectiveness of the similarity metric.} In response to inquiries about our choice of cosine similarity as the metric for assessing parameter similarities, we conducted comprehensive experiments comparing various similarity metrics, including cosine similarity, Euclidean distance, and Manhattan distance. The experimental results given in Table \ref{tab:metric_eval} reveal that cosine similarity and Euclidean distance consistently outperformed Manhattan distance. 
\begin{table}[htbp]
    \centering
    \scalebox{0.8}{
    \begin{tabular}{c|c|c|c|c}
    \toprule
         Similarity metric & MBPP   & CommonSenseQA  & GSM8K  & CMMLU  \\
    \hline
    Consine similarity  & 43.00 & 75.27    & 70.05 & 48.41\\
    Euclidean Distance  & 43.00 & 75.01   & 70.09 & 48.30 \\
    Manhattan Distance  & 42.85 & 74.86   & 69.12 & 48.40 \\
    \bottomrule
    \end{tabular}}
    \caption{The comparison results between different similarity metrics.}
    \label{tab:metric_eval}
\end{table}

\noindent\textbf{Pruning under different pruning ratios.} As mentioned in \textbf{Implementation}, our approach enhances the existing channel merging methods. Previous implementations of channel merging, such as TIES and DARE, necessitate pruning the delta parameters before merging to reduce parameter conflicts. Accordingly, we adopt the same strategy of pruning delta parameters prior to merging. We conducted experiments with various pruning ratios to determine their impact on the final results. Our findings indicate that a 30\% pruning ratio yields the most effective outcome. The specific experimental results demonstrating this are presented in Table \ref{tab:ratio_eval}.
\begin{table}[htbp]
    \centering
    \scalebox{0.83}{
    \begin{tabular}{c|c|c|c|c|c}
    \toprule
         Ratio & MBPP   & HumanEval & MMLU   & GSM8K  & CMMLU  \\
    \hline
    0\%   & 42.06 & 43.98 & 61.70 & 68.36 & 47.38\\
    30\%   & 43.00 & 45.13 & 63.01 & 70.05 & 48.41 \\
    50\%   & 43.00 & 45.01 & 62.80 & 70.09 & 48.40 \\
    70\%   & 42.80 & 44.69 & 62.53 & 69.15 & 48.27 \\
    \bottomrule
    \end{tabular}}
    \caption{The comparison results between different pruning ratios.}
    \label{tab:ratio_eval}
\end{table}


\end{document}